\documentclass{article}
\usepackage{spconf,amsmath,graphicx}
\usepackage{booktabs,tabularx}
\usepackage{amssymb,bm}

\title{DARTS: Dialectal Arabic Transcription System}
\twoauthors
 {Sameer Khurana, James Glass}
	{MIT\\
	Computer Science \& Artificial Intelligence Laboratory\\
	Cambridge, MA, USA}
 {Ahmed Ali}
	{HBKU\\
	Qatar Computing Research Institute\\
	Doha, Qatar}

\begin{document}
%
\maketitle
\begin{abstract}
We present the speech to text transcription system, called DARTS, for low resource Egyptian Arabic dialect. We analyze the following; transfer learning from high resource broadcast domain to low-resource dialectal domain and semi-supervised learning where we use in-domain unlabeled audio data collected from YouTube. Key features of our system are: A deep neural network acoustic model that consists of a front end Convolutional Neural Network (CNN) followed by several layers of Time Delayed Neural Network (TDNN) and Long-Short Term Memory Recurrent Neural Network (LSTM); sequence discriminative training of the acoustic model; n-gram and recurrent neural network language model for decoding and N-best list rescoring. We show that a simple transfer learning method can achieve good results. The results are further improved by using unlabeled data from YouTube in a semi-supervised setup. Various systems are combined to give the final system that achieves the lowest word error on on the community standard Egyptian-Arabic speech dataset (MGB-3).
\end{abstract}
\section{Introduction}
\label{sec:intro}

There are a number of major challenges associated with multi-dialect automatic speech recognition (ASR) of conversational Multi-Genre Broadcast (MGB) media, such as background noise variation, cross-talk, and transcriber inconsistency of reference transcripts~\cite{ali2018word}. Dialectal Arabic speech recognition, especially, suffers from the lack of enough in domain transcribed data~\cite{ali2018multi}. There have been many previous attempts to reduce the word error rate (WER) in MGB domain. Both English ASR MGB-1 ~\cite{bell2015mgb} and the first editon of the Arabic ASR in MGB-2 ~\cite{ali2016mgb} focused on using  mainstream broadcast media (BBC in MGB-1, Al Jazeera in MGB-2). The MGB-3 ~\cite{ali2017speech} used YouTube recordings to extend the diversity of the challenge and deal with dialectal Arabic as a typical example of languages which do not have well-defined orthographic rules and not enough transcribed data.

Deep learning has shown great benefit from big data. For example, the MGB-2 best system ~\cite{ khurana2016qcri} achieved more than 7\% relative improvement in WER  by augmenting the speech data from $1,200$ to $3,600$ as proposed in ~\cite{ ko2015audio}. 


In this work, we use a simple transfer learning technique combined with semi-supervised learning \cite{manohar2018semi} using unlabelled YouTube data and a very deep neural network based acoustic model to achieve state-of-the-art performance on the challenging low-resource Egyptian Arabic Dialectal speech recognition.

\section{Corpus Description}
We use the Arabic MGB-2 \cite{ali2016mgb} and MGB-3 \cite{ali2017speech} challenge dataset for developing the speech transcription system.
The MGB-2 challenge data consists of $1,200$ hours of Arabic broadcast news programs from $2005$-$2015$ made available by AlJazeera news channel. The programs are transcribed manually, not in a verbatim fashion. The transcription contains para-phrasing and other  issues. The quality of the transcription varies. AlJazeera programs that 
were used to develop the dataset which consisted of 10 years of recordings across 19 program series. The recorded programs can be categorized into three broad classes; a) conversations, where the program host talks with more than one guest discussing current affairs, b) interviews, where the program host interacts with a single guest, and c) report, such as news and documentaries. Conversational speech is particularly challenging due to overlapping speech and presence of various dialects. The MGB-2 data can be further broken up into 12 topics. More information about the dataset can be found in \cite{ali2016mgb}.
To construct the final training dataset, the audio was transcribed using a lightly supervised decoding process, that gives timing information for the audio in the dataset. The output of the decoding process for each spoken utterance was compared to the ground truth transcription and the error rate between the two is calculated at the word level (Matched Error Rate) and at the grapheme level (Phone Matched Error Rate). The Matched WER and Phone Matched WER along with Average Word Duration threshold was used to select reasonable accurate speech segments for training the acoustic models.

Unlike the MGB-2 dataset, in which most acoustic segments are spoken in Modern Standard Arabic (MSA) 
, the MGB-3 dataset is majority Egytian (EGY) Dialect. The MGB-3 dataset consists of only 16 hours of transcribed speech collected from YouTube, which makes the problem of accurate transcription quite challenging. The 16 hours of data is collected across seven topics: comedy, family, fashion, drama, sports, and TEDx. The data is split into development, adaptation and evaluation. We combine the adaptation and development data and use it for training the acoustic models.

Along with the labeled $1,200$ hours MGB-2 and 16 hours MGB-3 dataset, we have more than $500$ hours of unlabeled YouTube audio data collected from Egyptian Arabic YouTube channels. As the unlabeled data is from the same domain as our target MGB-3 dataset, using it in a semi-supervised setup could give us better performance. To collect the YouTube data, we manually select YouTube channels that are from the part of the world where majority of people speak Egyptian dialect. For this we use our knowledge of the Arabic Language and its Egyptian dialect. Once we have identified the desired YouTube channels, we search for videos with certain keywords that lets us extract videos of varied genres such as cooking, comedy, news and sports. Most of the videos collected are recorded by daily "YouTubers" talking informally in their local dialect. We further ran the state-of-the-art Arabic dialect classifier that is trained to predict the five Arabic dialects, Egyptian being one of them \cite{shon2018convolutional}.

In this work, we use a simple transfer learning technique that combines the MGB-2 and MGB-3 datasets that are used for training the acoustic models. We do not perform any data cleaning and use the data as given.

A large corpus of background text is also available for building language models which consists of over $130$ million words crawled from the AlJazeera website AlJazeera.net from $2000$--$2011$. We use the BuckWalter \footnote{Buckwalter is a one-to-one mapping allowing non Arabic speakers to understand Arabic scripts, and it is also left-to-right, making it easy to render on most devices.} format for the transcriptions, provided by the previous years MGB-2 and MGB-3 challenge organizers. The transcripts do not contain any punctuation or dicraization and
we also did not use any text surface orthographic normalization for three characters; \textit{Alef}, \textit{yaa}, \textit{taa} and \textit{marbouta}, which are often mistakenly written in dialectal text. This normalization is standard for dialectal Arabic pre-processing and reduces the sparseness in the text.
The language model is represented as a finite state transducer (FST) which is composed with grammar and lexicon FST to build the final decoding graph. For training and decoding the acoustic models we use a Grapheme lexicon i.e. the discrete units that the acoustic model is trained to classify are the language graphemes instead of phonemes. The major reason to use a grapheme lexicon is that it requires no linguistic knowledge and can be easily extracted from the text available in the speech database. This makes it easy to build accurate systems in cases where appropriate linguistic knowledge is not available. Admittedly, there is some noise injected in the training process while using a grapheme lexicon but one could argue that the noise can act as a regularizer to prevent overfitting to a particular dialect. Given enough data, the performance of a grapheme and a phoneme based speech transcription system is almost similar \cite{wang2018phonetic}. We use this corpus to construct \textit{n}-gram and Recurrent Neural Network language models which are used for re-scoring the decoded lattices.

\section{DARTS: Dialectal Arabic Transcription System}
\begin{figure*}
\includegraphics[width=\linewidth]{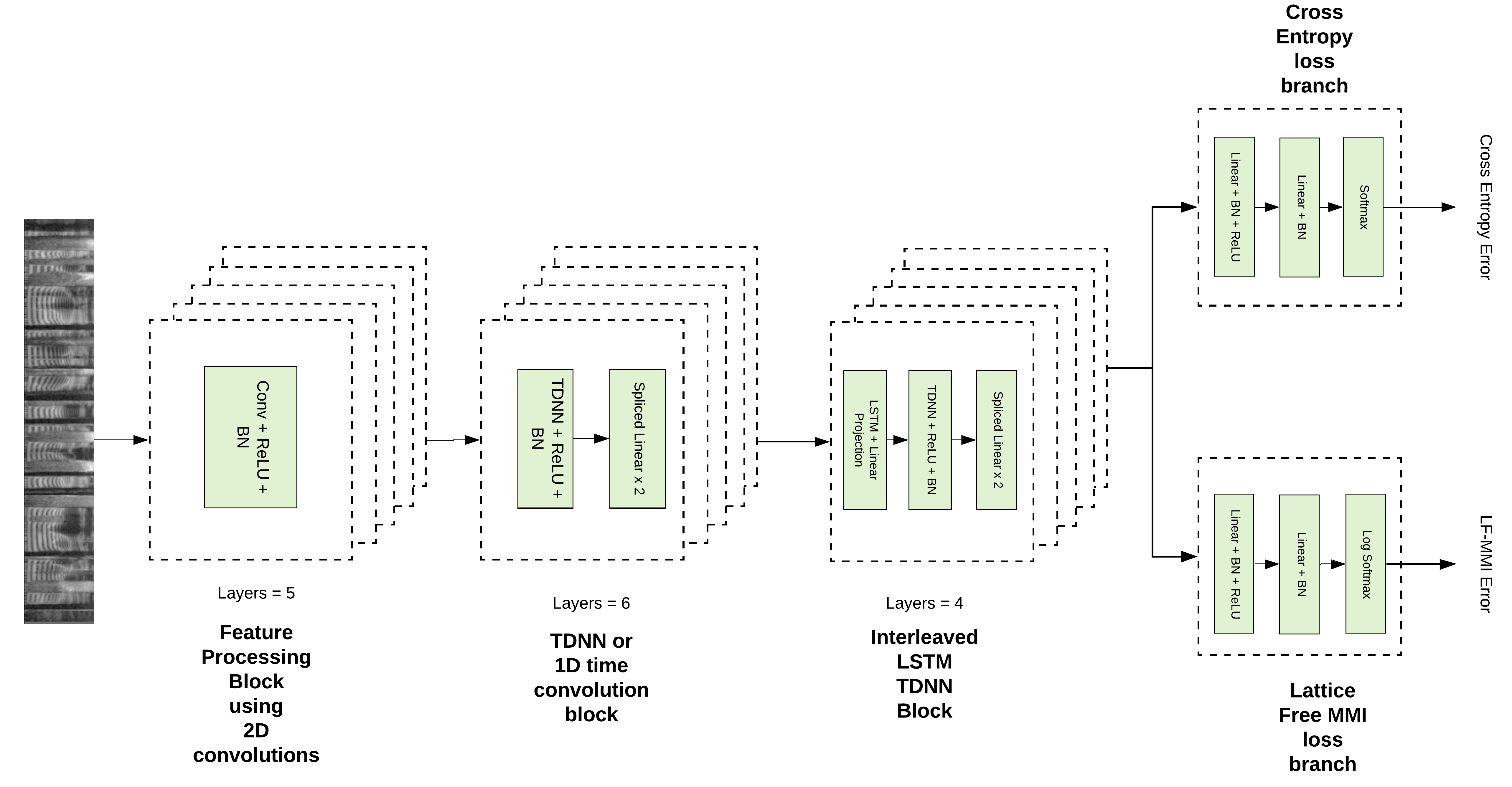}
\caption{Acoustic Model Architecture.}
\label{fig:1}
\end{figure*}
In this section, we give a high level conceptual overview of the various components of our speech-to-text transcription system. This sections also acts as background needed to understand the speech transcription system. The following sections contain more specific details about the various moving parts that make up DARTS.

Although, end-to-end systems \cite{amodei2016deep,kim2017joint,miao2015eesen,pratap2018w2l} are all the rage, our speech to text transcription system is a Hybrid HMM-DNN system \cite{swietojanski2013revisiting}. Hybrid systems are still the state-of-art systems on various datasets, they require less amount of training data and no linguistic knowledge is necessary for constructing the language specific lexicon because grapheme based systems perform as well, if not better than the phoneme based systems \cite{wang2018phonetic}.

The two main components of the speech transcription system are acoustic models and language models. Acoustic models are tasked with learning a robust mapping from continuous speech signals to the discrete acoustic units. The target acoustic units are generally tri-phones that are provided by a baseline Hidden-Markov Model (HMM) with a Gaussian Mixture Model as the likelihood function. The states of the HMM correspond to the acoustic units while the GMM is used to model continuous acoustic frames. The acoustic frames are extracted from the raw speech waveform by performing Cepstral Analysis that gives the commonly used Mel-Frequency Cepstral Coefficients (MFCCs) \cite{sigurdsson2006mel}. The baseline system HMM-GMM system, trained using the Forward-Backward Maximum Likelihood framework, is used to align the acoustic frames of a speech utterance with HMM states (each state correspond to an acoustic unit) to generate the training data examples, $(x, y)$, for training the Deep Neural Network acoustic model, where $x \in \mathbb{R}^{d}$ is the acoustic frame and $y$ is the HMM state (\textit{one of k}). Traditionally, the DNN acoustic models have been trained using the Cross Entropy (CE) training criterion. 

The CE training criterion with Softmax output layer pushes up the probability of the correct label and pulls down the probability of the competing labels. CE is a frame-based training criterion which ignores the fact that speech transcription is a sequence labeling problem. Sequence level training criterion based on Maximizing the Mutual Information (MMI) between the predicted  and the ground truth label sequence has been proposed for the DNN acoustic models \cite{vesely2013sequence}. MMI training criterion is given by the following equation:
\begin{eqnarray*} \label{eq:1}
\mathcal{L} &=&  \sum\limits_{u=1}^{U} log p_{\lambda}(M(w_{u}) | \bm{x_u}) \\
&=& \sum\limits_{u=1}^{U} log \frac{p_{\lambda}(\bm{x_u} | M(w_{u})) p(w_u)}{\sum\limits_{w} (\bm{x}_u | M(w))p(w)}
\end{eqnarray*}
For training utterances $\bm{x_1, x_2, \ldots, x_u}$ where $w_u$ is the word transcription for the $u^{th}$ utterance, $M(w_u)$ is the corresponding HMM and $\lambda$ is the set of HMM parameters. Some End-to-End speech recognition systems map $\bm{x_u}$ directly to words, $w_u$, but in our work the mapping problem is from $\bm{x_u}$ to HMM states, $M(w_u)$. Numerator of equation \ref{eq:1} corresponds to likelihood of the HMM model clamped to reference alignment, $w_u$, while denominator is the normalization that makes $\mathcal{L}$ a valid probability distribution. Intuitively, the MMI training criterion is maximized when the probability of the correct word sequence (numerator) is pushed up while for the rest of sequences (denominator) the probability is pulled down, hence, giving a sequence level training criterion. Weighted finite state transducers (WFST) lattices are used to compute equation \ref{eq:1} efficiently.

To train a DNN with training criterion \ref{eq:1}, \cite{vesely2013sequence} proposed to initially train the DNN with CE error function. The DNN is then used to generate alignment lattices for sequence training. Once the numerator and denominator lattices are in order, re-train the DNN with the sequence level criterion \ref{eq:1}. Sequence trained systems performs significantly better than CE trained systems \cite{vesely2013sequence}. 

\cite{povey2016purely} improved upon the previous work by introducing Lattice-Free MMI (LF-MMI) training criterion, which obviates the need for pre-computing training lattices using a CE trained DNN model and compute lattices on the fly. LF-MMI training criterion when used alone can be unstable for training very deep neural network acoustic models and hence, is used in conjunction with CE training criterion in a multi-task setup (Multi-LF-MMI).  Multi-LF-MMI based acoustic models achieve the state-of-the-art performance on various benchmarks. Practical implementation of these methods is a whole different ball-game, but fortunately an efficient implementation exists in \textit{kaldi speech recognition toolkit} \cite{povey2011kaldi}. A semi-supervised version of LF-MMI was introduced in \cite{manohar2018semi}. Mathematically, it requires a small tweak the LF-MMI equation presented earlier in this section. The semi-supervised LF-MMI is given by the following equation:
\begin{equation}
\mathcal{L} = \sum\limits_{u=1}^{U}\frac{\sum\limits_{w\in \mathcal{H}}P(x_u| M(w)) P(w)}{\sum\limits_{w^{\prime}}P(x_u| M(w^{\prime})) P(w^{\prime}) }
\end{equation}
Notice, the difference in the numerator term between the LF-MMI equation and the Semi-supervised LF-MMI equation. Because, we do not know the ground truth transcription for the utterance $x_{u}$, therefore the numerator now becomes the sum over possible hypotheses $\mathcal{H}$. To tractable compute the semi-sup LF-MMI objective function, \cite{manohar2018semi} proposed to first train a seed model trained on the available labeled data and then generate possible hypotheses for a given utterance by decoding it using the seed model. The decoding process given a bunch of hypotheses which can then be used to compute the semi-supervised objective. This is a fairly simple method to make use of unlabled data to train the acoustic model. Although, the method is theoretically easy to understand, it is not easy to implement. For more practical details reader is directed to the paper \cite{manohar2018semi}

Multi-LF-MMI and the semi-supervised version of Multi-LF-MMI form the backbone of our work. For more details readers are directed to the following excellent works \cite{povey2016purely}

Language models (LMs), another essential component of a speech transcription system, are used for beam search decoding and re-scoring hypothesis. For first-pass beam search decoding, usually smaller n-gram LMs are used to give a list of possible word sequences. The list is then re-scored using larger n-gram and RNN LMs to give the final predictions.

More concretely, in our work we experiment with very deep neural network acoustic models that consist of Time Delayed Neural Networks (TDNNs) \cite{peddinti2015time}, Convolutional Neural Networks (CNNs) \cite{lecun1995convolutional} and Long Short-Term Memory Recurrent Neural Networks (LSTMs) \cite{hochreiter1997long} as their building blocks. The input to our acoustic models is 
a high resolution MFCCs \cite{sarikaya2000high}, concatenated with 100 dimensional speaker specific i-vectors that facilitates speaker adaptation. We use Multi-LF-MMI training criterion to train the acoustic models on speech-to-acoustic units alignments generated using a baseline HMM-GMM system. The baseline HMM-GMM system is a K state Hidden Markov Model with state specific Gaussian Mixture Model likelihood function. The states of the HMM correspond the context dependent phones. In our case the context size is three. The system is trained using the Baum Welch algorithm. After training, the HMM is used to provide frame to state alignments using the viterbi algorithm. The Deep Neural Network is then trained on these alignments. 

We use tri-gram language model built using training transcripts for first-pass beam search decoding and four-gram and RNN LMs trained on the background text and training transcripts for re-scoring the decoded hypotheses.

The training data is augmented using speech and volume perturbation techniques. In previous work \cite{khurana2016qcri}, they found data augmentation to be extremely crucial for getting state-of-the-art results and hence, in this work we go with data augmentation as a default setting.
In next sections, we give concrete details about acoustic model and language model configurations and summarize the results.
\section{Acoustic Model}
\label{sec:majhead}

\begin{table}
\center
\begin{tabular}{lllr}
\toprule
ID & Layer & offset & dim\\
\midrule
   1l & TDNN &  \{-1, 0, 1\} & 1280 \\
   2l & Linear  & \{-1, 0\}  & 256 \\
   3l & Linear& \{-1, 0\}  & 256 \\
   4l & TDNN &  \{0, 1\}   & 1280 \\
   5l & Linear &  \{-1, 0\} & 256 \\
   6l & Linear &  \{-1, 0\} & 256 \\
   7l & TDNN &  \{0, 1\} & 1280\\
   8l   & Linear & \{-1, 0\}  & 256\\
   9l & Linear & \{-1, 0\}  & 256\\
   10l & TDNN & \{0, 6l, 3l\} & 1280 \\
   11l & Linear & \{-3, 0\} & 256 \\
   12l & Linear & \{-3, 0\} & 256 \\
   13l & TDNN & \{0, 3\} & 1280\\
   14l & Linear &   \{-3, 0\} & 256\\
   15l & Linear &  \{0, 3\} & 256\\
   16l & TDNN & \{0, 9l, 6l, 3l\} & 1280\\
   17l & Linear & \{-3, 0\} & 256\\
   18l & Linear & \{0, 3\} & 256\\
\bottomrule
\end{tabular}
\caption{DARTS TDNN Block. The model contains residual connections to make the training of the deep acoustic model stable. The repetition pattern can also be inferred from this table.}
\label{tab:1}
\end{table}

In Figure~\ref{fig:1}, we give a pedagogical view of our final acoustic model. The model is similar to deep speech \cite{amodei2016deep} in some dimensions and different in others. Similar to deep speech, our model uses $2$D Convolutional Neural Networks (CNN) layers for feature pre-processing. Unlike deep speech the CNN layers in our model are followed by Time Delayed Neural Network (TDNN) (or $1$D time convolutions) layers which are followed by Interleaved Recurrent and TDNN layers. For recurrence we use LSTM cell. Finally, we have two parallel loss function branches, one for Cross Entropy error function and the other for LF-MMI error function. Below we give details about each individual block in the acoustic model.

\begin{table}
\center
\begin{tabular}{lllr}
\toprule
ID & Layer & offset & dim \\
\midrule
   1r & LSTMp &  - & 1024,256,128 \\
   2r & TDNN  & \{0, 3\}  & 1280 \\
   3r & Linear& \{-3, 0\}  & 256 \\
   4r & Linear &  \{-3, 0\}   & 256 \\
   5r   & LSTMp & -  & 1024,256,128\\
   6r & TDNN & \{0, 15l, 12l, 9l\}  & 256\\
   7r & Linear & \{0, 3\} & 1280 \\
   8r & Linear & \{-3, 0\} & 256 \\
   9r & LSTMp &  - & 1024,256,128\\
   10r & TDNN & \{0, 8r, 4r, 18l\} & 1280\\
   11r & Linear & \{-3, 0\} & 256\\
   12r & Linear & \{0, 3\} & 256\\
   13r & LSTMp &  \{0, 3\} & 1024,256,128\\
\bottomrule

\end{tabular}
\caption{DARTS Recurrent Block. the three comma separated values in the dim column for LSTMp is the cell dimension, hidden dimension and linear projection dimension. This block also consists of residual connections, some coming in from the DARTS TDNN block, such as $15$l, $12$l, $9$l and $18$l.}
\label{tab:2}
\end{table}

\begin{table}
\center
\begin{tabular}{lllllr}
\toprule
Model & Tokens & Vocab & Vocab-LM & PPL \\
\midrule
  3-gram  & 9M & 300k & 1M & 611 \\
  4-gram & 133M & 1.5M & 1M & 436  \\
  RNN   & " & " & 80k & 481 \\
\bottomrule
\end{tabular}
\caption{Language Models. Token is the total number of words in the corpus. Vocab is the unique tokens. LM-vocab is the restricted vocab for the LM. RNN LM has a much smaller vocab than n-gram. This is due to the GPU memory restrictions.}

\label{tab:lm}
\end{table}

\subsection{Building Blocks}
 \subsubsection{DARTS Front-End}
CNN layers used in our acoustic model are standard two dimensional convolutions followed by ReLU non-linearity and Batch Normalization. The DARTS acoustic model has $5$ CNN layers acting as front-end feature processing. DARTS CNN front end is given below:
\begin{figure}
\includegraphics[width=80mm]{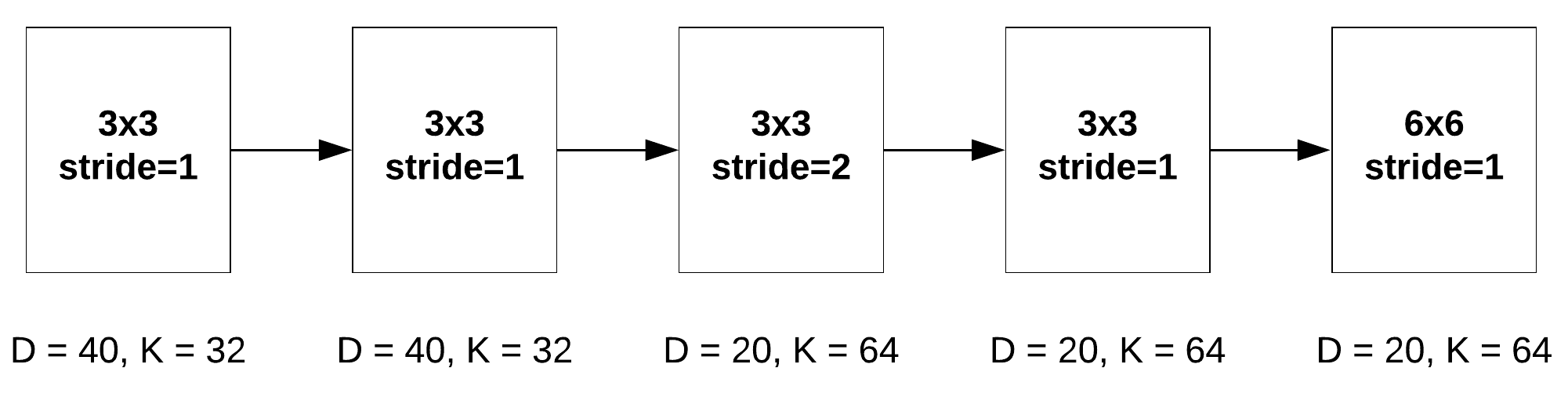}
\caption{DARTS CNN Front End. Consists of two Convolution operations acting on the input speech signal having feature dimension of $40$ and width equal to the number of acoustic frames that make up the sequence. The first two CNN blocks are $3x3$ convolutions with stride=1 and number of filters=$32$. Followed by a subsampling CNN layer with stride=$2$, followed by two final CNN layers. K corresponds to the number of filters and F corresponds to the input feature dimension.}
\end{figure}
 
 \subsubsection{DARTS TDNN}
DARTS CNN front end is followed by $6$ TDNN computation layers. Each  TDNN computation layer consists of a TDNN layer followed by $2$ spliced linear layers. As the name implies, spliced linear layer takes as input current time step from the below layer spliced with the time step given by an offset value.

TDNN is a popular computational layer used in speech transcription systems. TDNN is just one dimensional convolution along the time axis of a time series signal. A popular variant of TDNN is the sub-sampled TDNN presented in \cite{peddinti2015time}, that takes an input spliced  feature vectors from the below layers. The splicing indexes are given by the offset values. For example, the offset configuration for a TDNN layer given by $\{-1, 0\}$ implies that input feature sequence to the TDNN layer is the current feature vector spliced with vector to the left of the current time step. For more details reader is refereed to \cite{peddinti2015time}. In our case, the offset configurations are given in the table \ref{tab:1}.

\subsubsection{DARTS Recurrent}
  Following the DARTS TDNN block is the DARTS Recurrent block. This block consists of the following repeating pattern: LSTMp $\rightarrow$ TDNN $\rightarrow$ Linear $\rightarrow$ Linear $\rightarrow$. The pattern is repeated four times in the DARTS Recurrent block. The LSTMp layer is the LSTM layer followed by linear projection, with cell dimension of 1024, hidden dimension of 256 and linear projection layer of dimension 128. The TDNN layer is one dimensional time convolution followed by ReLU non-linearity and Batch Normalization. The full configuration with TDNN offsets is given in table \ref{tab:2}.

\section{Language Models}
We do not use the background text provided by the organizers of the MGB-2 challenge for the baseline HMM-GMM system. We use the MGB-2 and MGB-3 training trascripts combined to develop a tri-gram language model developed using the SRILM toolkit. We use the Kneser-Ney (KN) method as the smoothing algorithm for the tri-gram language model. The language model was further pruned to reduce its size. The language model is converted to a Finite State Transducer (FST) that is used to make the decoding graph for the first pass decoding. First pass decoding gives us decoding lattices that contain plausible output hypothesis. The list of hypothesis are then rescored using a four gram language model. The four gram language model is developed by using the combined MGB-2, MGB-3 and the background text. The background text consists of more than $130$ Million tokens. The four gram language model also undergoes KN smoothing and pruning of the rare N-grams.

Furthermore, we developed a Recurrent neural network language model (RNN-LM) for rescoring the list of hypothesis. For developing the RNN-LM we used \textit{mikolov's} RNN language modeling toolkit \cite{mikolov2012statistical}. In this work, we train an RNN LM with MaxEnt connections (RNNME). RNNME consists of both recurrent connections and non-recurrent connections or direct connections between the input and output layer. These are the MaxEnt connections or the Maximum Entropy connections. Essentially, this model jointly trains a recurrent and an \textit{n}-gram language model and has been shown to perform better than just RNNLM. Due to a large vocabulary size, we train a class based RNNME LM with the following hyper-parameter settings; $200$ classes, $40$K most frequent words as the vocabulary size and also the input size of the neural network, $300$ neurons in the RNN hidden layer, sigmoid activation function and $2000$M direct connections between the input and output layer with the n-gram order of three. We acknowledge that today we have better language models such transformers \cite{devlin2018bert} etc. and we leave the detailed language modeling investigation for the future work.

\section{Results}
In table \ref{tab:3} we show the WER comparisons of various models on the MGB-3 eval set. Details about the eval set can be found in \cite{ali2017speech}. MGB2+3 data setting is the simple transfer learning setup where we first train the acoustic model on $1,200$ hours of MGB-2 broadcast news data which is in Modern Standard Arabic, then we re-train the model on $16$ hours of MGB-3 Egyptian dialect data. The final model is evaluated on MGB-3 eval set that was used for evaluation in the MGB-3 challenge \cite{ali2017speech}. This setup is exactly same as the challenge.

The key takeaway from the results table is that CNN pre-processing block helped in improving the performance. DARTS model with LSTM layers perform better than the BLSTM layers which is due to overfitting on MGB-2 dataset. The DARTS BLSTM model gives a WER of $17$\% on MGB-2 development set compared to $18.5$\% WER given by DARTS LSTM model, which shows overfitting on MGB-2 in case of DARTS BLSTM.

The YouTube data used in the semi-supervised setup \cite{manohar2018semi} helps in improving the ASR system performance. The system combination gives the best result which is further improved by $4$-gram and RNNLM language model re-scoring. We achieve the state-of-the-art result on the MGB-3 challenge dataset using the DARTS system which improves upon the previous state-of-the-art result of $37.5$\% which consisted of a system combination of $30$ different systems \cite{smit2017aalto}.

To wrap up our work, we performed another set of experiments on the recently introduced Moroccan Arabic dialect as part of the MGB-5 challenge \cite{ali2019mgb5}. The training data consisted of labelled $10$ hours collected from $69$ YouTube programs of varied genres, the development data consisted of $2$ hours collected from $10$ YouTube programs and the unseen test data was also $2$ hours. Following the same transfer learning strategy we trained the DARTS system on the combined MGB-2 and MGB-5 datasets. We achieved significantly better results than the baseline TDNN network that was trained only on the MGB5 training data $65$\% vs $70$\%. The transfer from MGB-2 to MGB-5 is not as drastic as for MGB-2 to MGB-3. This is because MGB-2 dataset which is spoken in Modern Standard Arabic dialect is linguistically close to spoken Egyptian than to Moroccan dialect. Moroccan dialect being from the North African region is closer to the French language. In the future, it would be interesting to quantify linguistic similarity between languages and choose the closest language to perform transfer learning. 

 \begin{table}
\center
\begin{tabular}{llllr}
\toprule
ID & Model & Data & MGB-3 \\
\midrule
  1 & \small{TDNN} & MGB2+3 & 42.2 \\
  2 & TDNN-LSTM & MGB2+3 & 40.1  \\
  3 & TDNN-BLSTM  & MGB2+3 & 41.9  \\
  4 & CNN-TDNN  & MGB2+3 & 41.8  \\
  5 & DARTS-BLSTM  & MGB2+3 & 41.5  \\
  5 & DARTS  & MGB2+3 & 39.8  \\
  6 & CNN-TDNN  & MGB2+3+YT & 39.11  \\
  7 & DARTS & MGB2+3+YT &  39.5  \\
  8 & DARTS-BLSTM & MGB2+3+YT &  40.5  \\
  9 & 5+6+7+8 & - & 36.5  \\
  10 & 5+6+7+8 + LM Rescoring & - &  \textbf{35.8}  \\
\bottomrule

\end{tabular}
\caption{WER\% Results on MGB-2 and MGB-3 datasets with different acoustic models. Tri-gram LM is used for first-pass decoding. YT refers to unlabeled YouTube data. DARTS is the single best system. System combination of various models gives the best performance. LM re-scoring improves the result further and achieve the new state-of-the-art improving upon the previous best of $37.5$.}
\label{tab:3}
\end{table}

\section{Conclusions}
In this work, we used a simple transfer learning technique combined with semi-supervised learning and very deep acoustic models to achieve state-of-the-art performance on the challenging low-resource dialectal Arabic speech recognition (MGB-3). In the future, we will explore end-to-end speech recognition systems and explore YouTube data collection strategies to develop a more robust speech recognition that can be used for real world applications. It would also be interesting to explore challenges associated with developing ASR systems for other Arabic dialects.
\bibliographystyle{IEEEbib}
\bibliography{main}

\end{document}